\documentclass[journal]{IEEEtran}
\usepackage{graphicx}
\usepackage{bm} % bold symbol
\usepackage{multirow}
\usepackage{colortbl}
\usepackage{tikz}
\newcounter{example}

\newcommand{\ignore}[1]{}
\newcommand\copyrighttext{%
  \footnotesize 
  \copyright \copyright 20xx IEEE. Personal use of this material is permitted. Permission from IEEE must be obtained for all other uses, in any current or future media, including reprinting/republishing this material for advertising or promotional purposes, creating new collective works, for resale or redistribution to servers or lists, or reuse of any copyrighted component of this work in other works.\\
  This work has been submitted to the IEEE for possible publication. Copyright may be transferred without notice, after which this version may no longer be accessible.}
\newcommand\copyrightnotice{%
\begin{tikzpicture}[remember picture,overlay]
\node[anchor=north,yshift=-3pt] at (current page.north) {\fbox{\parbox{\dimexpr\textwidth-\fboxsep-\fboxrule\relax}{\copyrighttext}}};
\end{tikzpicture}%
}
\begin{document}

\title{Constructing Multilayer Perceptrons as Piecewise Low-Order Polynomial Approximators: \\
A Signal Processing Approach}

\author{Ruiyuan~Lin, Suya~You, Raghuveer~Rao and C.-C.~Jay Kuo % <-this % stops a space
\thanks{Ruiyuan~Lin and C.-C.~Jay~Kuo are with Ming Hsieh Department
of Electrical and Computer Engineering, University of Southern
California, 3740 McClintock Avenue, Los Angeles, USA,
emails:\{ruiyuanl,jckuo\}@usc.edu}% <-this % stops a space
\thanks{Suya You and Raghuveer Rao are with Army Research Laboratory,
Adelphi, Maryland, USA, emails:\{suya.you.civ,raghuveer.m.rao.civ\}@mail.mil}%
}% 
% The paper headers 
%\markboth{IEEE Signal Processing Letters,~Vol.~x, No.~x, Month~Year}% 
%{Lin \MakeLowercase{\textit{et al.}}: Relationship of Two Approximations } 

\maketitle

\copyrightnotice

\begin{abstract}

The construction of a multilayer perceptron (MLP) as a piecewise
low-order polynomial approximator using a signal processing approach is
presented in this work. The constructed MLP contains one input, one
intermediate and one output layers. Its construction includes the
specification of neuron numbers and all filter weights.  Through the construction, a one-to-one correspondence
between the approximation of an MLP and that of a piecewise low-order
polynomial is established. Comparison between piecewise polynomial and
MLP approximations is made. Since the approximation capability of
piecewise low-order polynomials is well understood, our findings shed
light on the universal approximation capability of an MLP. 

\end{abstract}


\IEEEpeerreviewmaketitle

\section{Introduction}\label{sec:introduction}

Piecewise low-order polynomial approximation (or regression) is a
classic topic in numerical analysis. It fits multiple low-order
polynomials to a function in partitioned intervals. The approximation
capability of piecewise low-order polynomials is well understood.  In
contrast, the representation power of neural networks is not as obvious.
The representation power of neural networks has been studied for three
decades, and universal approximation theorems in various forms have been
proved \cite{cybenko1989approximation, hornik1989multilayer,
chui1992approximation, chen1995universal, scarselli1998universal}. These
proofs indicate that neural networks with appropriate activations can
represent a wide range of functions of interest.  Their goals were to
make their theorems as general as possible. The proofs were based on
tools from functional analysis (e.g., the Hahn-Banach theorem
\cite{rudin1973functional}) and real analysis (e.g., Sprecher-Kolmogorov
theorem \cite{sprecher1965structure, kolmogorov1957representation}),
which is less well known in the signal processing community.
Furthermore, although they show the existence of a solution, most of
them do not offer a specific MLP design. Lack of MLP design examples
hinders our understanding of the behavior of neural networks. 

A feedforward construction of an MLP with two intermediate layers was
presented to solve a classification problem in \cite{lin2020two}. As a
sequel to \cite{lin2020two}, this work addresses the regression problem.
For ease of presentation, our treatment will focus on the approximation
of a univariate function.  We adopt a signal processing (or numerical
analysis) approach to tackle it.  Through the design, we can explain how
MLPs offer universal approximations intuitively.  The construction
includes the choice of activation functions, neuron numbers, and filter
weights between layers.  A one-to-one correspondence between an MLP and
a piecewise low-order polynomial will be established. Except for the
trivial case of the unit step activation function, results on MLP
construction and its equivalence to piecewise low-order polynomial
approximators are new.  Another interesting observation is the critical
role played by activation functions. That is, activation functions
actually define kernels of piecewise low-order polynomials.  Since the
approximation error of piecewise low-order polynomials is well
understood, our findings shed light on the universal approximation
capability of an MLP as well. 

The rest of this paper is organized as follows.  The construction of
MLPs corresponding to pieceswise constant, linear and cubic polynomial
approximations is detailed in Sec.  \ref{sec:construction}.  The
generalization from the 1-D case to multivariate vector functions is
sketched in Sec. \ref{sec:generalization}.  Concluding remarks and
future research directions are given in Sec.  \ref{sec:conclusion}. 

\section{Proposed MLP Constructions}\label{sec:construction}

Without loss of generality, we consider a continuous function $f(x)$
defined on interval $x\in[0,1]$, which is uniformly divided into $N$
sub-intervals. Let $h=N^{-1}$ be the length of the subinterval and
$x_i=ih$, where $i=0,1,\cdots,N$. We show how to construct an MLP so
that it serves as a piecewise constant or a piecewise linear
approximation to $f(x)$ in this section. 

\subsection{Piecewise Constant Approximation}\label{subsec:constant}

A piecewise constant approximation of $f(x)$ can be written as
\begin{equation}\label{eq:piecewise_constant}
f(x) \approx f_{c} (x) = \sum_{i=0}^{N-1} f(x_i) b_i (x),
\end{equation}
where 
\begin{equation}\label{eq:box}
b_i (x) = \left\{
\begin{array}{ll}
1, \quad & x_i \le x < x_{i+1} \\
0, \quad & \mbox{otherwise}.
\end{array}
\right.
\end{equation}
is the unit box function. We can rewrite the unit box function
as the difference of two unit step functions
\begin{equation}\label{eq:box-2}
b_i (x) = u (x-x_i) - u (x - x_{i+1}), 
\end{equation}
where 
\begin{equation}\label{eq:step}
u(x) = \left\{
\begin{array}{ll}
1, \quad & x \geq 0 \\
0, \quad & \mbox{otherwise},
\end{array}
\right.
\end{equation}
is the unit step function. Then, we can rewrite Eq. (\ref{eq:piecewise_constant}) 
as
\begin{equation}\label{eq:piecewise_constant_2}
f_{c} (x) = f(x_0) u(x-x_0) + \sum_{i=1}^{N-1} [f(x_{i})-f(x_{i-1})] u (h^{-1}(x-x_i)).
\end{equation}

Our designed MLP consists of one input node denoting $x\in[0,1]$, one
output node denoting $f(x)$, and $N$ intermediate nodes (or neurons),
denoted by $R_j$, where $j=0,1,\cdots,N-1$.  We use $\alpha_j$ and
$\xi_j$ to denote the weight and bias between the input node, $R_{in}$
and $R_j$ and $\tilde{\alpha}_j$ and $\tilde{\xi}_j$ to denote the
weight and bias between $R_j$ and the output node $R_{out}$,
respectively. The response at $R_j$ before activation, denoted by $y_j$, and 
the response at $R_{out}$, denoted by $z$, can be written as
\begin{eqnarray}
y_j & = & \alpha_j x + \xi_j, \label{eq:y_j} \\
z & = & \sum_{j=0}^{N-1} \tilde{\alpha}_j \mbox{Act} (y_j) + \tilde{\xi}_j, \label{eq:z} 
\end{eqnarray}
where Act is the activation function, $\alpha_j$ and $\tilde{\alpha}_j$ are weights
and $\xi_j$ and $\tilde{\xi}_j$ are biases.

With the unit step function as activation and Eq.
(\ref{eq:piecewise_constant_2}), it is easy to verify that we can choose
the following weights and biases for the MLP:
\begin{equation}\label{eq:weight} 
\begin{array}{lll}
R_{in} \Rightarrow R_j: & \alpha_j=h^{-1}, & \xi_j= - h^{-1} x_j, \\
R_j \Rightarrow R_{out}: & \tilde{\alpha}_j=f(x_{j})-f(x_{j-1}), & \tilde{\xi}_j=0,
\end{array}
\end{equation} 
where $f(x_{-1})\equiv 0$. The above derivation was given in
\cite{scarselli1998universal}, and it is repeated here for the sake of
completeness. 

\subsection{Piecewise Linear Approximation}\label{subsec:linear}

A commonly used nonlinear activation function is the rectified linear
unit (ReLU), which can be written as
\begin{equation}\label{eq:relu}
r(x)=\max (0,x).
\end{equation}
We will construct an MLP using $r(x)$ as the nonlinear activation
function and show its equivalence to a piecewise linear
approximation to $f(x)$. The latter can be expressed as
\begin{equation}\label{eq:piecewise_linear}
f(x) \approx f_{l} (x) = \sum_{i=0}^{N-1} f(x_i) t_i (x),
\end{equation}
where 
\begin{equation}\label{eq:triangle-2}
t_i (x) = t(h^{-1}(x-x_i))
\end{equation}
and where $t(x)$ is the unit triangle function in form of
\begin{equation}\label{eq:triangle}
t(x) = \left\{
\begin{array}{ll}
x+1,  & \mbox{if } -1 \leq x \leq 0 \\
-x+1,  & \mbox{if } \; 0 \leq x \leq 1 \\
0,    & \mbox{otherwise}. 
\end{array}
\right.
\end{equation}

\begin{figure}[th]
\begin{center}
\includegraphics[width=0.6\linewidth]{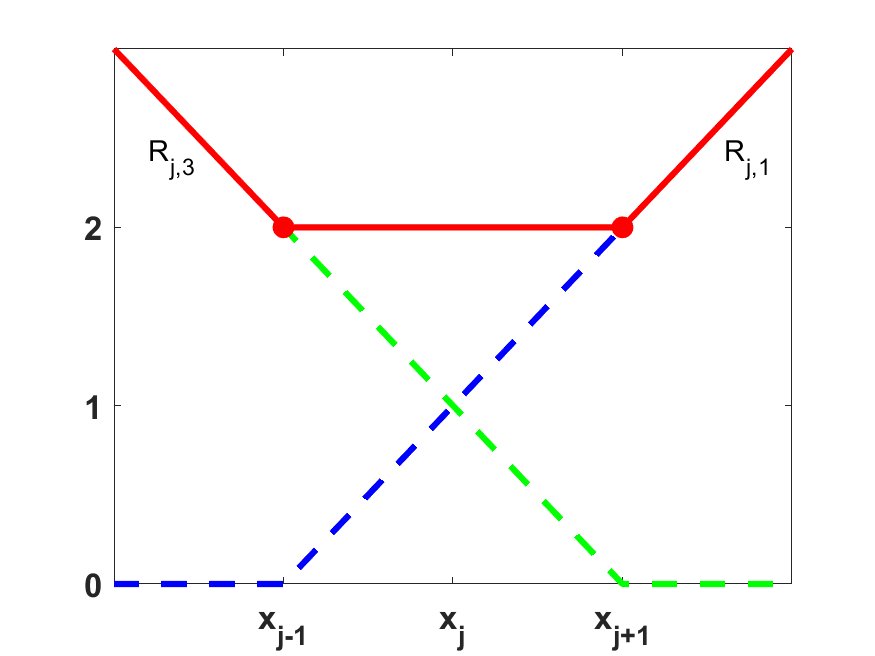} \\
(a) \\
\includegraphics[width=0.6\linewidth]{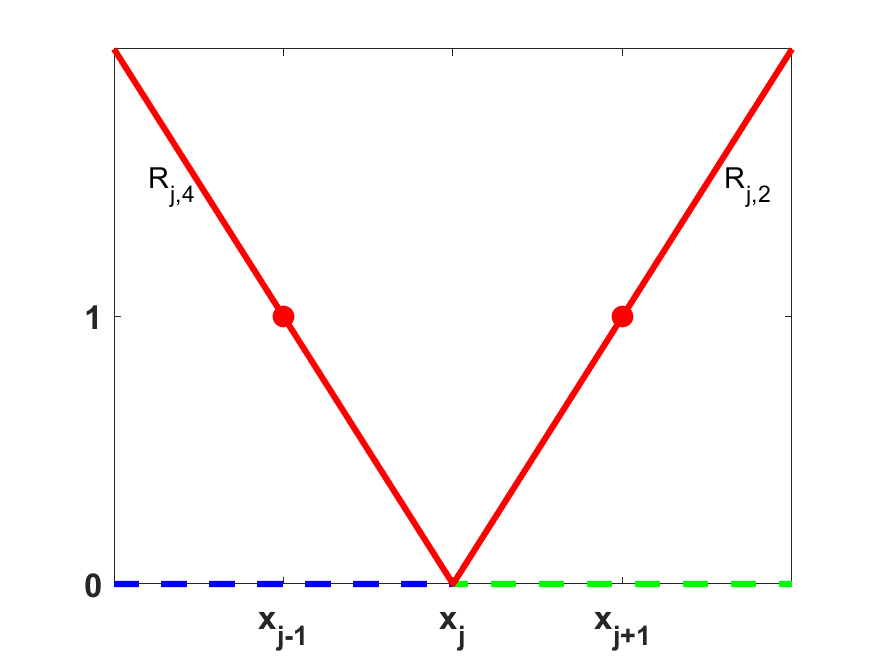} \\
(b) \\
\includegraphics[width=0.6\linewidth]{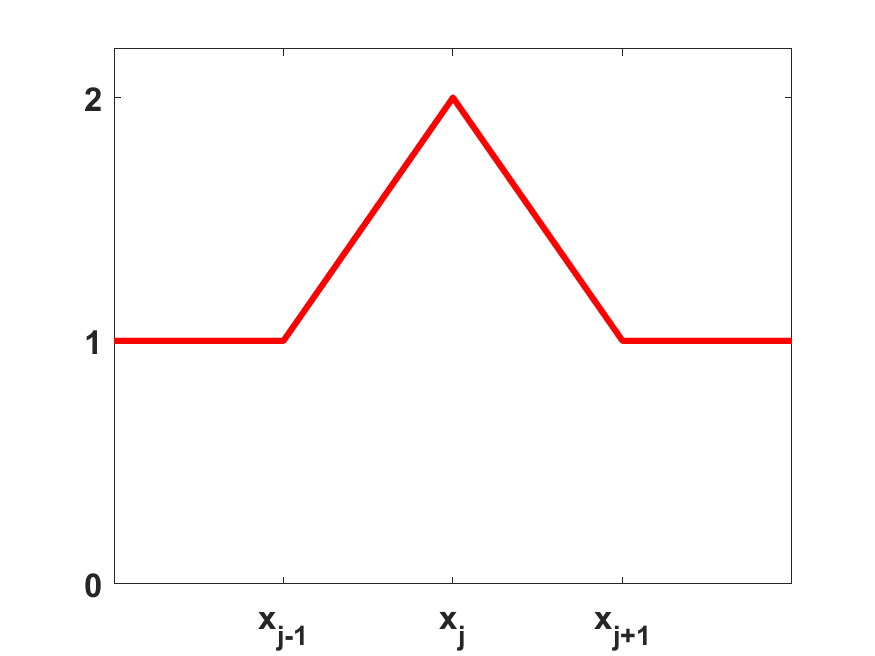} \\
(c) \\
\includegraphics[width=0.6\linewidth]{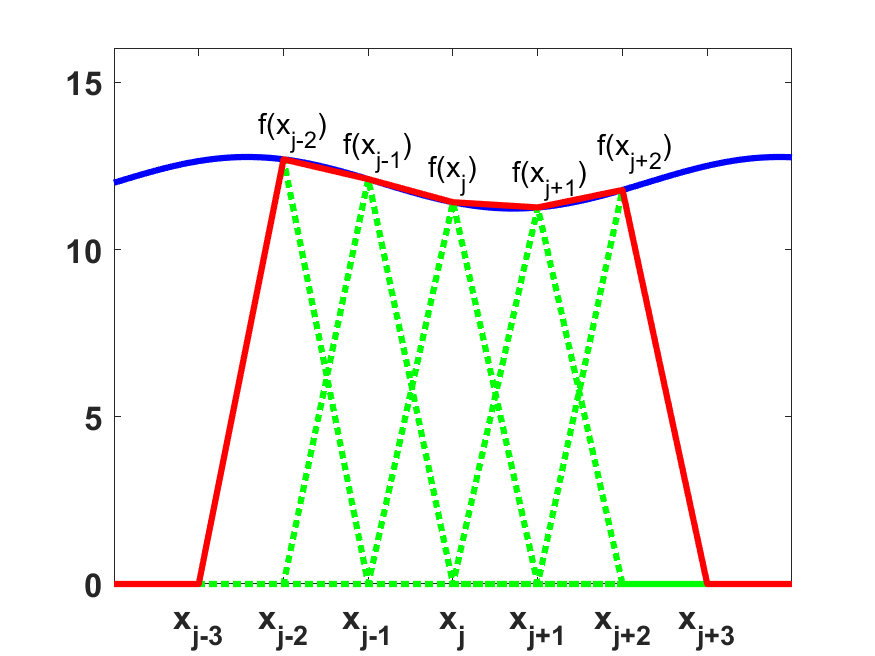} \\
(d)
\end{center}
\caption{Responses of (a) neurons $R_{j,1}$ and 
$R_{j,2}$ individually (dotted lines) and jointly (solid lines), (b)
neurons $R_{j,3}$ and $R_{j,4}$, (c) all four neurons combined, and
(d) the approximation of $f(x)$ with a sequence of weighted triangle
functions.}\label{fig:1}
\end{figure}

With one input node denoting $x\in[0,1]$ and one output node denoting
$f(x)$, we construct an MLP that has 4 neurons in two pairs, denoted by
$R_{j,k}$, $k=1,\cdots,4$, in the intermediate layer to yield the same
effect of $t_j$.  We use $\alpha_{j,k}$ and $\xi_{j,k}$ to denote the
weight and bias from the input node $R_{in}$ to $R_{j,k}$, respectively.
In our design, they are specified as:
\begin{equation}\label{eq:pl-1}
\begin{array}{lll}
R_{in} \Rightarrow R_{j,1}: & \alpha_{j,1} =  h^{-1}, & \xi_{j,1} = - h^{-1} x_{j-1}, \\
R_{in} \Rightarrow R_{j,2}: & \alpha_{j,2} = h^{-1},  & \xi_{j,2} = - h^{-1} x_j, \\
R_{in} \Rightarrow R_{j,3}: & \alpha_{j,3} = - h^{-1}, & \xi_{j,3} = h^{-1} x_{j+1}, \\
R_{in} \Rightarrow R_{j,4}: & \alpha_{j,4} = - h^{-1}, & \xi_{j,4} = h^{-1} x_j. \\
\end{array}
\end{equation}
The responses of neurons $R_{j,k}$, $k=1,3$ and $k=2,4$ after ReLU
nonlinear activation are shown in Figs. \ref{fig:1} (a) and (b),
respectively, where the dotted lines indicate those of individual
neurons and the solid lines show those of combined responses. Finally,
the combination of all responses of all 4 neurons at the output node is
shown in Fig. \ref{fig:1} (c). It is a unit triangle plus a vertical
shift. 

%%%%%%%%%%%%%%%%%%%%%%%%%%%%%%%%%%%%%%%%%%%%%%%%%%
%%%%%%%%%%%%%%%%%%%%%%%%%%%%%%%%%%%%%%%%%%%%%%%%%%

Similarly, we use $\tilde{\alpha}_{j,k}$ and $\tilde{\xi}_{j,k}$ to
denote the weight and bias from $R_{j,k}$ to the output node $R_{out}$,
respectively. They are specified below:
\begin{equation}\label{eq:pl-2}
\begin{array}{lll}
R_{j,k} \Rightarrow R_{out}: & \tilde{\alpha}_{j,k} = f(x_j), & k=1,3 \\
R_{j,k} \Rightarrow R_{out}: & \tilde{\alpha}_{j,k} = - f(x_j), & k=2,4 \\
R_{j,k} \Rightarrow R_{out}: & \tilde{\xi}_{j,k}= - 0.25 f(x_j), & 1\le k \le 4.
\end{array}
\end{equation}
It is easy to verify from the Fig. \ref{fig:1} (d) that these four neurons are nothing
but a weighted triangle centered at $x_i$ with its base between
$x_{i-1}$ and $x_{i+1}$ and height $f(x_i)$ and $f(x)$ can be
approximated by a sequence of such weighted triangles. 

\subsection{Piecewise Cubic Approximation}\label{subsec:cubic}

\begin{figure}[th]
\begin{center}
\includegraphics[width=0.6\linewidth]{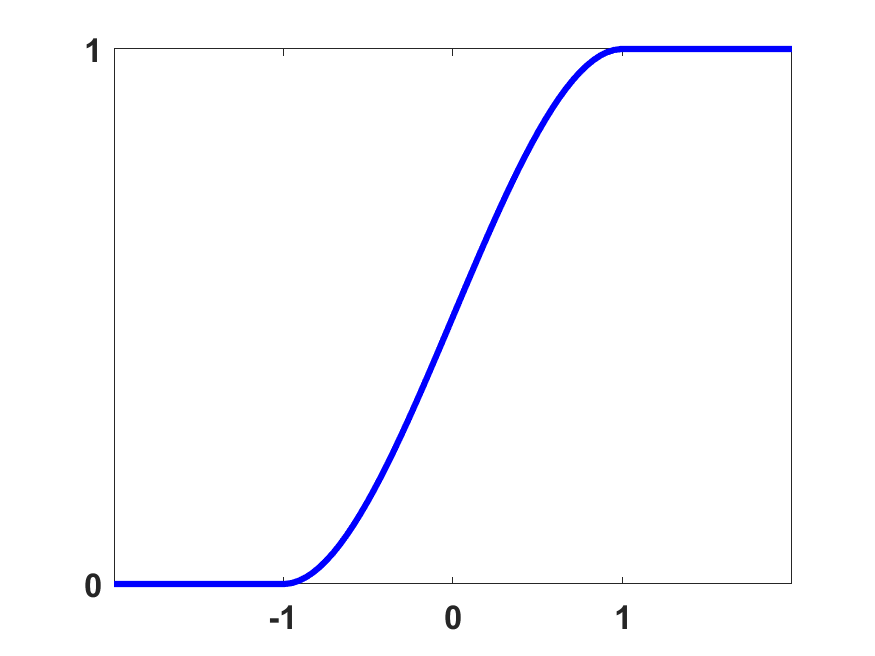} \\
(a) \\
\includegraphics[width=0.6\linewidth]{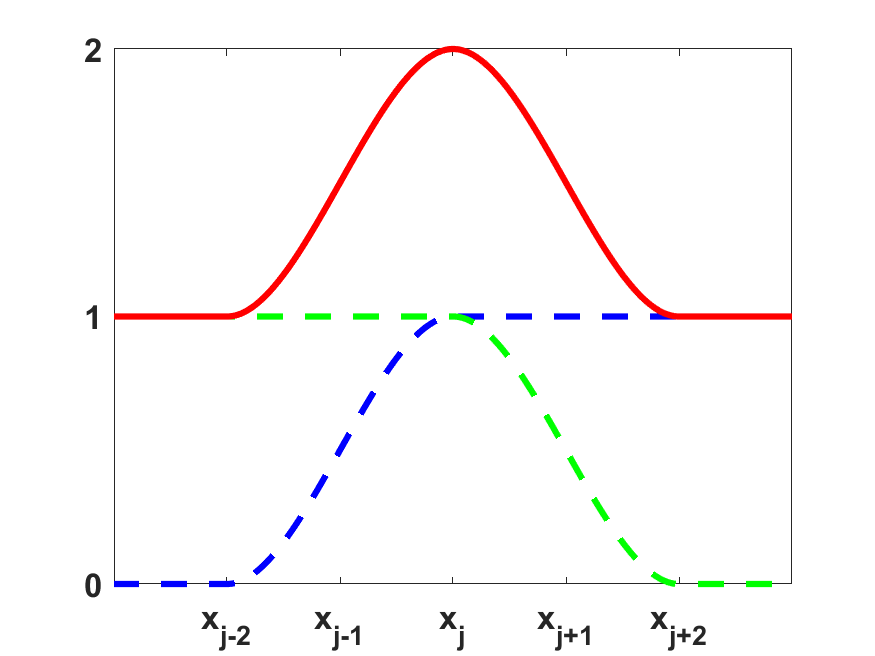} \\
(b) \\
\includegraphics[width=0.6\linewidth]{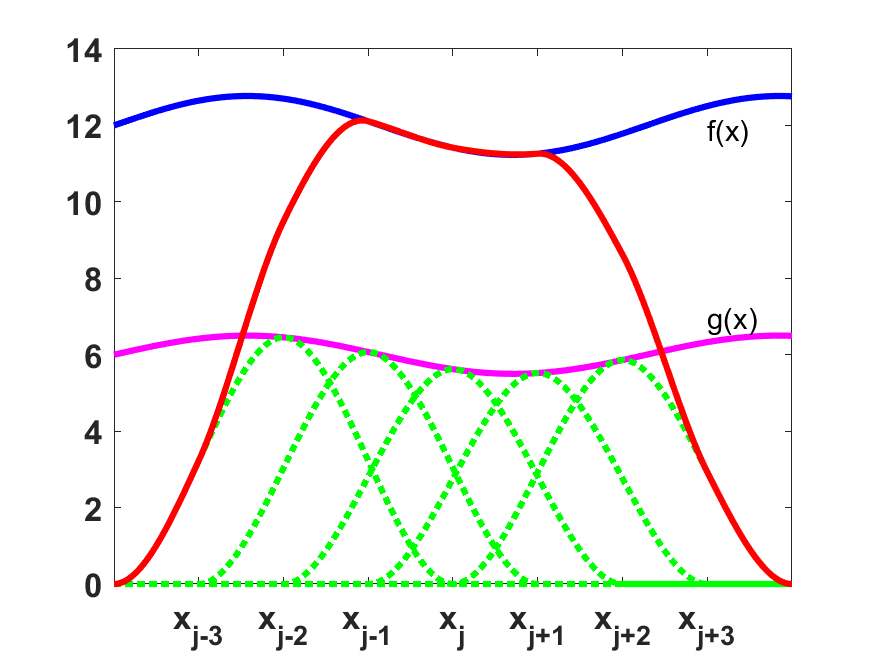} \\
(c) \\
\includegraphics[width=0.6\linewidth]{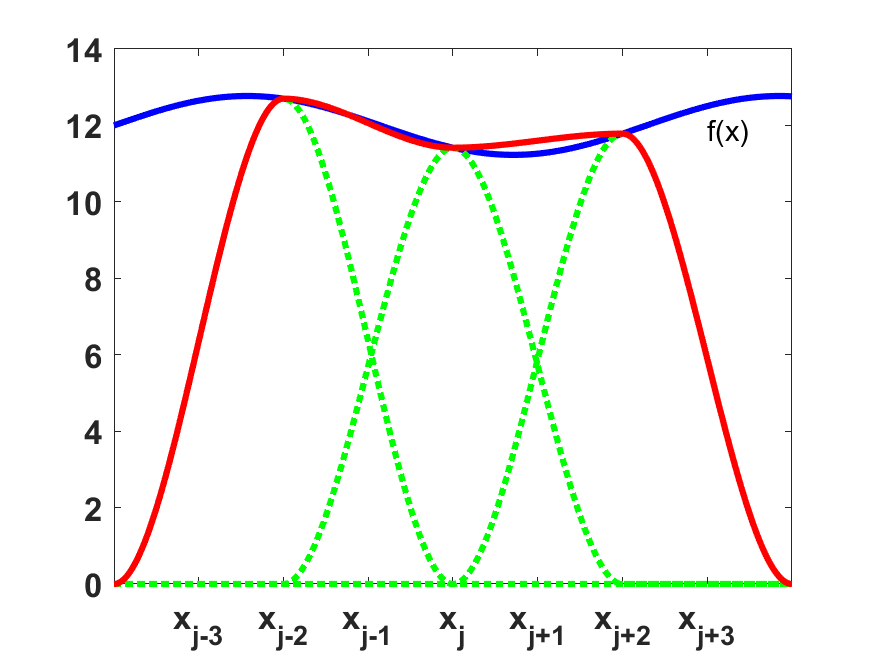} \\
(d)
\end{center}
\caption{(a) The unit cubic activation function, (b) responses of
$R_{j,1}$ and $R_{j,2}$ individually (dotted lines) and jointly (solid
lines), (c) and (d) two designs for the approximation to $f(x)$ with a
sequence of weighted third-order bumps.}\label{fig:2}
\end{figure}

Next, we design an MLP that offers a piecewise cubic approximation to
$f(x)$. Besides incoming and outgoing weights/biases, we will design
the unit cubic activation (see Fig. \ref{fig:2}(a)) in form of: 
\begin{equation}\label{eq:cubic}
q(x)=\left\{
\begin{array}{ll}
0 & x \le -1, \\
a_3 x^3 + a_2 x^2 + a_1 x + a_0, \; & -1 \le x \le 1, \\
1 & x \ge 1, 
\end{array}
\right.
\end{equation}
where $q(x)$, $-1 \le x \le 1$, satisfies two constraints: i) $q(-1)=0$
and, ii) $q(x)$ is anti-symmetric with respect to $(0,0.5)$; i.e.,
\begin{equation}\label{eq:q1}
q(x)+q(-x)=1.
\end{equation}
Note that these two constraints imply $q(0)=0.5$ and $q(1)=1$.  By
substituting $q(-1)=0$, $q(0) =0.5$, $q(1)=1$ into Eq. (\ref{eq:cubic}),
we get $a_2=0$, $a_1+a_3=0.5$, and $a_0=0.5$. There is one degree of
freedom left. To complete the design, one choice is to specify the
first-order derivative at the inflection point: $(x,f(x))=(0,0.5)$.  As
shown in Fig. \ref{fig:2}(a), the inflection point has the maximum
first-order derivative and its second-order derivative is zero. 

%%%%%%%%%%%%%%%%%%%%%%%%%%%%%%%%%%%%%%%%%%%%%%%%%%
%%%%%%%%%%%%%%%%%%%%%%%%%%%%%%%%%%%%%%%%%%%%%%%%%%

For interval $[x_{j-2},x_{j+2}]$ centered at $x_j$, we can use two
neurons to build a third-order unit bump function plus a vertical shift
as shown in Fig. \ref{fig:2}(b).  The weight and bias from the input
node $R_{in}$ to $R_{j,k}$ are specified as:
\begin{equation}\label{eq:q-1}
\begin{array}{lll}
R_{in} \Rightarrow R_{j,1}: & \alpha_{j,1} = h^{-1}, & \xi_{j,1} = - h^{-1} x_{j-1}, \\
R_{in} \Rightarrow R_{j,2}: & \alpha_{j,2} = - h^{-1},  & \xi_{j,2} = h^{-1} x_{j+1} \\
\end{array}
\end{equation}
where $R_{j,1}$ and $R_{j,2}$ are activated by the unit cubic function,
$q(x)$.  The weights and biases from $R_{j,k}$, $k=1,2$, to output node
$R_{out}$ are:
\begin{equation}\label{eq:q-2}
R_{j,k} \Rightarrow R_{out}: \; \tilde{\alpha}_{j,k} = g(x_j), \; 
\tilde{\xi}_{j,k}=-0.5 g(x_j), 
\end{equation}
where $g(x_j)$, $j=0,\cdots,N$, is the solution to the following linear 
system of $(N+1)$ equations:
\begin{equation}\label{eq:q-3}
g(x_j)+0.5[g(x_{j-1})+g(x_{j+1})]=f(x_j), \; 0 \le j \le N,
\end{equation}
with boundary conditions $g(x_{-1})=g(x_{N+1})=0$. Given the target
regression function $f(x_j)$ at the right-hand-side of (\ref{eq:q-3}),
one can solve the system for $g(x_j)$, $j=0,\cdots,N$, uniquely.
If $f(x)$ is smooth in a local region, $g(x) \approx 0.5 f(x)$
as shown in Fig. \ref{fig:2}(c).

There is an alternative design as shown in Fig. \ref{fig:2}(d), where
we increase the spacing between two adjacent bumps from $h$ to $2h$.
The weight and bias from the input node $R_{in}$ to $R_{j,k}$ still
follow Eq. (\ref{eq:q-1}) with $j=2j'$, where $j'=0,1,\cdots, N/2$,
while Eq. (\ref{eq:q-2}) should be changed to 
\begin{equation}\label{eq:q-4}
R_{j,k} \Rightarrow R_{out}: \; \tilde{\alpha}_{j,k} = f(x_j), \; 
\tilde{\xi}_{j,k}=-0.5 f(x_j).
\end{equation}
This is fine since there is no interference between $f(x_{j-2})$,
$f(x_j)$ $f(x_{j+2})$. Furthermore, using the Taylor series
expansion\footnote{ under the assumption that low-order derivatives
of $f(x)$ are bounded} and based on the fact $x_{j+1}$ is the inflection
point of two associated cubic activation functions, we can derive
\begin{equation}\label{eq:q-5}
0.5 [f(x_{j+2})+f(x_j)] - f(x_{j+1}) \approx O(h^4),
\end{equation}
The design in Fig. \ref{fig:2}(c) demands $2N$ neurons while that in
Fig. \ref{fig:2}(d) demands $N$ neurons only in total. The latter design
appears to be simpler.

\subsection{Discussion}\label{subsec:discussion}

This section builds a bridge between Piecewise low-order polynomial and
MLP approximations.  The approximation errors of piecewise low-order
polynomials can be derived using Taylor's series expansion. As
$N=h^{-1}$ goes to infinity, the errors of piecewise constant, linear
and cubic polynomials converge to zero at a rate of $O(h)$, $O(h^2)$ and
$O(h^4)$. The same conclusion applies to the corresponding designed MLPs.

\begin{figure}[th]
\begin{center}
\includegraphics[width=0.6\linewidth]{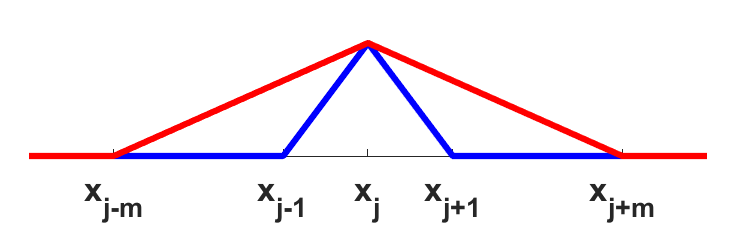}
\end{center}
\caption{Two kernels of different supports centered at $x_j$, which can be
achieved by adding more neurons to the intermediate layer of an MLP.} 
\label{fig:3}
\end{figure}

There is however difference between them. To give an example, consider
the kernel density estimation problem \cite{parzen1962estimation}:
\begin{equation}\label{eq:kernel-1}
\hat{f}_h(x)= \sum_{j=1}^N \omega_j K (h^{-1}(x-x_j)), 
\end{equation}
where $K(x)$ is a kernel.  The unknown function $f(x)$ at given point
$x$ can be expressed as the summation of weighted kernels, where weights
$\omega_j$ can be solved by a regression method (say, linear regression)
for given $K(x)$.  The box, triangle, and third-order unit bump
functions are kernels to smooth statistical variations in local
intervals.  The pre-selected kernel could be too local and rigid to be
efficient in handling long- and mid-range correlations in $f(x)$.  In
contrast, parameters in the first stage of an MLP allow a richer choice
of kernels.  Our MLP design does not exploit this flexibility fully. For
example, we may adopt more different $\alpha_{j,k}$ values  (see Fig.
\ref{fig:3}). 

%%%%%%%%%%%%%%%%%%%%%%%%%%%%%%%%%%%%%%%%%%%%%%%%%%
%%%%%%%%%%%%%%%%%%%%%%%%%%%%%%%%%%%%%%%%%%%%%%%%%%

One focal point of neural network research is to understand the role of
activation in a neuron. Historically, earliest activation was the simple
unit step function \cite{mcculloch1943logical}. It offers ``on" and
``off" two states with an abrupt jump. Such activation has zero
derivatives except at $x=0$ and, as a result, backpropagation is not
able to update parameters.  Several variants such as the sigmoid, ReLU,
leaky ReLU activations have been developed later. Paired activations are
used to build kernels of low-order polynomials here. As presented
earlier, higher-order nonlinear activations are not fundamental in the
universal approximation capability of neural networks. They affect the
kernel shape only. Instead, the role of nonlinear activation is
interpreted as a tool to resolve the sign confusion problem caused by
the cascade of multiple layers in \cite{kuo2016understanding} and
\cite{lin2020two}. 

\section{Generalization to Multivariate Vector Functions}
\label{sec:generalization}

Our presentation has focused on the case of 1D input/output so far.  In
general, the regression problem can be a $q$-dimensional vector function 
of $p$ multivariables:
\begin{equation}\label{generali}
F: {\bf x} \rightarrow {\bf f (x)}, \mbox{ where } {\bf x} \in [0,1]^p \; {\bf f} \in R^q,
\end{equation}
where $p,q \ge 1$. We will provide a sketch to the generalization from
the univariate scalar function to the multivariate vector function in
this section. Generalization of our proposed design to a vector function
is straightforward since we can design one MLP for each output dimention. 

The generalization from the univariate case to the multivariate case
using 1D piecewise low-order polynomials, we can proceed as follows:
\begin{enumerate}
\item Partition $[0,1]^p$ into $p$-D cells of volume size $h^p$;
\item Conduct the 1D approximation separately;
\item Perform the tensor product of 1D kernels to form $p$-D kernels.
\end{enumerate}
For the MLP design, we cannot perform the tensor product in the third
step directly. Instead, we can use the tensor product result to find the
weighted contributions of neighboring grid points to the center of each
cell and construct an MLP accordingly. 

In most practical machine learning applications, the multivariate vector
function, ${\bf f (x)}$, it is assumed to be sufficiently smooth.  That
is, its lower-order partial derivatives exist.  Consequently, Talyor
series expansion for multivariable functions can be adopted to estimate
the convergence rate to the target regression function. 

\section{Conclusion and Future Work}\label{sec:conclusion}

A bridge between piecewise low-order polynomials and three-layer MLP
approximators was built in this work.  Instead of following the
traditional path of universal approximation theorem of neural networks
based on functional or real analysis, we proposed a signal processing
approach to construct MLPs.  The construction includes the choice of
activation functions, neuron numbers, and filter weights between
adjacent layers. Our design decomposes an MLP into two stages: 1) kernel
construction in the first stage and 2) weight regression in the second
stage. We can explain how an MLP offers an universal approximation to a
univariate function intuitively by this approach.  Kernel design plays
an important role in many machine learning and statistical inference
problems. It was shown to be related to the shape of the activation
function here. 

Given an MLP architecture, backprogagation is used to solve kernel
construction and weight regression problems jointly via end-to-end
optimization. It is however an open problem to find the optimal number
of neurons in the intermediate layer.  On the other hand, the
traditional approach decouples the whole problem into two subproblems
and solve them one by one in a feedforward manner.  It is interesting to
study the relationship between the kernel design and the first-stage
filter weight/bias determination of MLPs.  It is also desired to exploit
prior knowledge about data and applications for flexible kernel design
in feedforward learning systems.

\newpage
\bibliographystyle{IEEEtran}
\bibliography{mybibfile}

\end{document}